# Artificial Intelligence and Accounting Research: A Framework and Agenda

November 2025

Theophanis C. Stratopoulos
School of Accounting and Finance, University of Waterloo
tstratopoulos@uwaterloo.ca

Victor Xiaoqi Wang
College of Business, California State University Long Beach
victor.wang@csulb.edu

**Abstract**: Recent advances in artificial intelligence, particularly generative AI (GenAI) and large language models (LLMs), are fundamentally transforming accounting research, creating both opportunities and competitive threats for scholars. This paper proposes a framework that classifies AI-accounting research along two dimensions: research focus (accounting-centric versus AI-centric) and methodological approach (AI-based versus traditional methods). We apply this framework to papers from the IJAIS special issue and recent AI-accounting research published in leading accounting journals to map existing studies and identify research opportunities. Using this same framework, we analyze how accounting researchers can leverage their expertise through strategic positioning and collaboration, revealing where accounting scholars' strengths create the most value. We further examine how GenAI and LLMs transform the research process itself, comparing the capabilities of human researchers and AI agents across the entire research workflow. This analysis reveals that while GenAI democratizes certain research capabilities, it simultaneously intensifies competition by raising expectations for higher-order contributions where human judgment, creativity, and theoretical depth remain valuable. These shifts call for reforming doctoral education to cultivate comparative advantages while building AI fluency.

**Keywords:** Artificial intelligence, generative AI, large language models, accounting research, research framework, doctoral education

---

We thank Efrim Boritz, Severin Grabski, Stewart Leech, Adam Presslee, and David Wood for their insightful comments. Professor Stratopoulos gratefully acknowledges financial support from the PwC Fund.

# Artificial Intelligence and Accounting Research: A Framework and Agenda

## 1. Introduction

Artificial intelligence is fundamentally transforming how accounting practitioners work, how investors and other users process financial information, and how accounting research is conducted. Practitioners are actively integrating these technologies into their work, realizing productivity and efficiency gains (Choi and Xie, 2025). The research community has responded similarly. Recent advancements in generative AI (GenAI) have generated substantial literature examining its impacts across accounting domains (Dong et al., 2024). Almost two-thirds of accounting researchers now use GenAI in their scholarly work (Barrick et al., 2025). Accounting journals, including the *International Journal of Accounting Information Systems* (IJAIS) and the *Journal of Accounting Research* (JAR), have issued calls for AI-accounting research.[1]

The seven papers in this IJAIS special issue exemplify the tremendous opportunities available in the field. Veganzones and Séverin (2025) apply machine learning to identify distinct earnings management strategies through financial profile analysis, demonstrating how AI can reveal patterns in accounting phenomena that traditional methods might miss. Hesse and Loy (2025) develop sentence-level BERT analysis of MD&A disclosures for bankruptcy prediction with enhanced interpretability, while Levich and Knust (2025) systematically compare discriminative versus generative models for multilingual accounting information extraction. Lin et al. (2025) convert financial ratios into images for CNN-based bankruptcy prediction and use SHAP and LIME to identify key ratios. Meanwhile, Blondeel et al. (2025) validate ChatGPT as a secondary coder in qualitative accounting research, and Fotoh & Mugwira (2025) conduct blind assessments revealing LLM capabilities and limitations in audit procedures. Dong et al. (2024) provide a comprehensive review of emerging literature on LLM applications across accounting and finance.

---

[1] IJAIS called for papers on "Artificial Intelligence, with a focus on Large Language Models, in Accounting," and published a special issue on this topic (https://www-sciencedirect-com.csulb.idm.oclc.org/special-issue/106TQXS22R1). JAR called for papers on "AI and Large Language Models in Accounting Research" for its 2025 conference (https://onlinelibrary-wiley-com.csulb.idm.oclc.org/doi/pdf/10.1111/1475-679X.12550).

These diverse contributions, spanning traditional accounting questions enhanced by AI, methodological innovations in AI itself, and critical evaluations of AI systems, illustrate the breadth of AI-accounting research. Yet their very diversity poses challenges: How should researchers, particularly those new to this intersection of literature, navigate this landscape? Where do these contributions fit relative to each other and to the broader literature? What types of expertise do different research approaches require? What research opportunities remain underdeveloped? And critically, where do accounting researchers' collaborative advantages lie relative to the capabilities that industry and computer science researchers bring to AI-related research in accounting?

These questions extend beyond the special issue to the rapidly growing AI-accounting literature more broadly. The literature is even more diverse and fragmented, as it spans multiple accounting subfields (financial reporting, auditing, taxation, management accounting, accounting information systems, and other specialized fields such as forensic accounting, sustainability reporting, and accounting education) and diverse methodological approaches (from machine learning algorithms to traditional empirical analysis and qualitative methods). While existing reviews have provided valuable insights through various analytical lenses (e.g., Abbas, 2025; Almufadda and Almezeini, 2022; Cao et al., 2024; Kassar and Jizi, 2025; Murphy et al., 2024; Ranta et al., 2023), they do not offer a comprehensive framework needed to simultaneously map existing literature and reveal strategic opportunities for future inquiry.

To address these questions, we propose a two-dimensional framework that classifies AI-accounting research according to its primary research focus (whether contributions advance accounting knowledge or AI capabilities for accounting applications) and methodological approach (whether studies employ AI-based methods or traditional research methods). This framework serves multiple purposes. First, it reveals where existing studies cluster and where opportunities remain underdeveloped. Second, it maps distinct technical skill requirements across research domains, helping scholars identify where their expertise can contribute most effectively and where collaboration might be beneficial. Third, it facilitates systematic evaluation of the field's evolution and identifies areas where collaboration between accounting, industry, and computer science researchers would be most valuable.

We demonstrate the framework's utility by applying it first to the seven papers in this special issue. This analysis shows that the special issue features strong contributions in methodological innovation, critical evaluation of AI capabilities, and AI-enhanced accounting inquiry. We then extend the analysis to classify 89 recent AI-accounting papers published in leading Accounting Information Systems (AIS) and general-interest (non-AIS) journals between 2022 and 2025. We start in 2022 to focus more on papers using recent generative AI technologies. We find that the distribution of AI-accounting research varies markedly between AIS and non-AIS journals, reflecting their distinct missions. AIS journals predominantly examine AI systems themselves through critical evaluation and methodological innovation, while non-AIS journals maintain focus on core accounting phenomena, deploying AI primarily as an advanced analytical tool. These contrasting patterns reveal different strategic positions: AIS journals exploring the technology frontier versus non-AIS journals integrating AI into established research paradigms.

These divergent patterns reflect journals' complementary missions rather than a hierarchy of research value. Indeed, both types of research should be welcomed and published in both AIS and non-AIS journals. AI-centric research provides essential methodological foundations and critical evaluations that enable and inform accounting-centric research. Similarly, accounting-centric research often grapples with technical challenges that motivate AI innovation. These bidirectional flows demonstrate that both research streams are necessary for the advancement of accounting research.

This systematic classification of the AI-accounting literature reveals not only where research has concentrated, but also where different research communities bring complementary strengths. The patterns we observe—particularly the divergence between AIS and non-AIS journals—reflect different but equally valuable approaches to AI-accounting research. These patterns matter because progress at the intersection of AI and accounting increasingly depends on combining the distinctive capabilities of accounting scholars, industry practitioners, and computer scientists. Recognizing these complementary strengths transforms our classification from a purely descriptive exercise into a framework for identifying productive collaborations.

Beyond organizing and synthesizing existing literature, we address a more fundamental question: How should accounting researchers strategically position themselves in an AI-transformed research landscape? The integration of AI into accounting creates both opportunities and collaborative imperatives. On one hand, AI provides powerful tools for idea generation, measurement, prediction, and analysis that can enhance traditional accounting inquiries and help researchers communicate insights effectively to different audiences. On the other hand, industry researchers possess superior computational resources and access to proprietary data, while computer science researchers maintain advantages in algorithmic innovation and technical sophistication.

While prior work has analyzed these dynamics through the lens of comparative advantage (Alles and Gray, 2025; Alles et al., 2008), we reframe this as collaborative advantage to recognize that progress at the intersection of AI and accounting benefits from the complementary strengths of different research communities. Through this collaborative advantage perspective, we identify research domains where accounting scholars' distinctive capabilities, such as deep institutional knowledge, theoretical grounding, causal inference expertise, and independent evaluation capacity, are most valuable. These capabilities position accounting scholars not only to make contributions that neither industry nor computer science researchers would naturally pursue, but also to collaborate effectively with industry and computer science researchers on problems that require both domain expertise and technical innovation.

We identify strategic opportunities in each research domain where accounting scholars' specialized expertise positions them for unique contributions. These range from developing specialized AI algorithms that prioritize transparency and regulatory compliance over proprietary advantage, to conducting independent critical evaluations that reveal AI's limitations and implementation challenges, to leveraging AI-generated measurements within rigorous causal inference designs for testing fundamental accounting theories, to exploiting AI-related events as natural experiments for studying core accounting phenomena.

We further recognize that AI's impact extends beyond research topics and methods to transform the research process itself fundamentally. Our analysis systematically compares human researchers and AI agents across the entire research workflow, from idea generation and literature review through data analysis and manuscript drafting, revealing their advantages. This comparison exposes a critical tension: while GenAI models can now assist with numerous research tasks, potentially democratizing access to specific capabilities, they simultaneously intensify competition by raising expectations for higher-order contributions where human judgment, creativity, and theoretical depth remain valuable, at least for the short term (i.e., 4 to 5 years). These shifts redistribute comparative advantages not only between academic, industry, and computer science researchers, but also across career stages within the accounting academic community itself. The implications are particularly acute for junior scholars and doctoral students, whose traditional roles as research apprentices are being disrupted by technologies that can now perform many tasks that once constituted their primary contributions to research teams.

In response to these challenges, we propose strategic responses at both the individual and institutional levels. For doctoral programs, we advocate adapting the traditional apprenticeship model for an AI-augmented environment, emphasizing the cultivation of skills where humans retain advantage, such as research design, causal inference, and theoretical integration, while building fluency in AI tools and validation protocols. For researchers across career stages, we outline positioning strategies that leverage accumulated expertise and established networks while adapting to AI-augmented workflows. The goal is not merely to survive this transition but to strategically exploit it, ensuring that accounting research maintains its intellectual rigor, practical relevance, and competitive vitality in an increasingly AI-intensive landscape.

The remainder of this paper proceeds as follows. Section 2 defines AI and delineates the scope of accounting used throughout the paper. Section 3 introduces our classification framework, demonstrates its application through detailed analysis of the IJAIS special issue papers, and then applies it to 89 papers published in leading accounting journals between 2022 and 2025. Section 4 identifies strategic research opportunities based on collaborative advantage analysis. Section 5 examines AI's role in transforming the research process with implications for doctoral student training. Section 6 provides concluding comments.

## 2. AI and Accounting

Artificial Intelligence (AI) encompasses systems capable of performing tasks that typically require human intelligence. Russell and Norvig (2020) define AI as "the study of agents that receive percepts from the environment and perform actions," identifying four fundamental approaches, namely, systems that think like humans, think rationally, act like humans, or act rationally. Contemporary policy frameworks, however, have converged on more operational definitions that emphasize AI's functional characteristics, particularly its role in information processing and decision support. The Organization for Economic Co-operation and Development (OECD, 2024) defines an AI system as "a machine-based system that, for explicit or implicit objectives, infers, from the input it receives, how to generate outputs such as predictions, content, recommendations, or decisions that can influence physical or virtual environments," while noting that such systems differ in their levels of autonomy and adaptiveness after deployment. Similarly, the EU Artificial Intelligence Act (2024) adopts a closely aligned definition, underscoring that AI systems are "designed to operate with varying levels of autonomy and that may exhibit adaptiveness after deployment."

Taken together, these converging definitions highlight three core characteristics of AI systems: (1) their ability to process inputs and generate outputs, (2) their capacity to shape decision-making environments, whether physical or virtual, and (3) their potential for autonomous and adaptive operation. This input-output decision-making framework parallels the fundamental structure of information systems, in which raw data is transformed into meaningful information to support organizational decision-making. This parallel becomes particularly salient when considering AI's integration into domain-specific systems such as accounting information systems.

Recent advances in artificial intelligence have given rise to Generative Artificial Intelligence (GenAI), a new branch of AI models capable of producing original content rather than merely classifying data or performing predictive analytics (U.S. Congressional Research Service, 2025). These systems, particularly those built on machine learning architectures such as large language models (LLMs), learn underlying patterns and structures from large training datasets to generate novel outputs, including text, images, videos, computer code, and music that resemble but do not replicate their training material (Feuerriegel et al., 2024).

Unlike traditional AI, which follows predetermined rules for narrow, task-specific problems, GenAI demonstrates broader reasoning and contextual understanding. These models have achieved remarkable capabilities, including answering accounting questions (Wood et al., 2023), passing professional certification exams such as the CPA and CMA (Eulerich et al., 2024), performing various tasks such as classification, summarization, and prediction at high proficiency (Wang and Wang, 2025), and extracting novel insights from large unstructured data (Brown et al., 2025). This paper covers both traditional and newer transformer-based AI technologies, with greater emphasis on the latter.

These AI capabilities are particularly well-suited to accounting, a discipline that emphasizes processing and interpreting large volumes of financial and non-financial data, summarizing complex transactions, performing analyses, and providing decision support for stakeholders. For this paper, we adopt a broad functional definition of accounting that encompasses its core disciplines: financial accounting, management accounting, auditing (both internal and external), taxation, accounting information systems (AIS), and other specialized areas (Coyne et al., 2010). This expansive scope reflects the growing integration of AI across diverse accounting contexts, from financial reporting to performance management (Libby and Presslee, 2025).

We further extend the scope of accounting to include the broader ecosystem of accounting information users, covering not only the traditional preparers and attestors of financial statements, but also investors who rely on accounting data for capital allocation decisions, financial analysts who interpret and evaluate corporate performance, creditors assessing creditworthiness, regulators ensuring market efficiency, and corporate executives who depend on accounting systems for strategic decision-making. This expansive view recognizes accounting as both a technical discipline and a critical information infrastructure that enables economic decision-making across multiple stakeholder groups, making it particularly relevant when examining AI's transformative potential across the entire accounting information chain. This broad view also aligns with the diverse range of topics one may find published in general interest accounting journals, which regularly feature research spanning these various functions and stakeholder perspectives.

## 3. Review of Recent AI-Accounting Research

### 3.1 AI-Accounting Research Framework

The rapid growth of AI research in accounting has produced an extensive and diverse literature spanning a wide range of research topics and methodological approaches. While existing reviews have provided valuable insights through various analytical lenses and focal points (e.g., Abbas, 2025; Almufadda and Almezeini, 2022; Cao et al., 2024; Kassar and Jizi, 2025; Murphy et al., 2024; Ranta et al., 2023), the field still lacks a comprehensive framework for organizing this growing literature. Such a framework would serve several important purposes.

First, it would enable researchers, particularly those new to this area, to quickly identify the core topics and understand various ways that they may contribute to this field. Second, it would provide a tool to systematically evaluate how existing studies advance our knowledge and reveal where gaps remain. Third, it would chart future research directions by identifying emerging trends, methodological innovations, and convergences across studies, thus helping researchers anticipate where the field is heading and position their work accordingly. Additionally, this framework would serve as a valuable tool for identifying where academic researchers maintain collaborative advantages (both relative to each other and to industry researchers and computer science researchers), thus enabling them to strategically position themselves within specific research domains and forge collaborations that leverage complementary expertise. To address this gap, we propose a framework that classifies AI-accounting research along two key dimensions: primary research focus and methodological approach.

Our classification system employs two dimensions: subject centricity and primary method. Subject centricity identifies where a paper's primary contribution lies, i.e., whether it advances accounting knowledge or AI capabilities for accounting applications. Research is classified as accounting-centric when its primary contribution enhances our understanding of various accounting domains, such as financial accounting, corporate disclosure, auditing, performance management, taxation, and governance practices. These studies treat accounting phenomena as their core focus, examining outcomes such as financial reporting quality, audit quality, internal control weaknesses, information processing, and market consequences. The defining characteristic is that accounting outcomes serve as the primary dependent variables of interest.

Conversely, studies are classified as AI-centric when their primary contribution advances AI methods, systems, or human-AI interaction within accounting contexts. These papers prioritize technical innovation and typically report model performance metrics as their primary outcomes, e.g., accuracy, precision, recall, F1 scores, AUC, and explainability indices. While these studies employ accounting data or settings, their core objective is methodological advancement or understanding the technology and its implementation, rather than generating accounting insights.

The methodological dimension distinguishes between AI-based and traditional methods. Research employs AI-based methods when AI tools or their outputs drive the main findings. These AI tools include LLMs, transformers, embedding models, topic models, neural networks, and machine learning algorithms. In contrast, research uses traditional methods when the key analysis relies on established research approaches, such as regression analysis, experiments, questionnaires, interviews, case studies, and literature reviews. Importantly, the mere use of AI for data preparation, e.g., optical character recognition, translation, data deduplication, or other preprocessing tasks, does not qualify a study as employing AI-based methods. The critical distinction is whether AI drives the core analysis and findings, not whether it facilitates data collection or preparation. Based on these dimensions, we propose the following 2×2 framework, as shown in Table 1.

### 3.1.1 Accounting-Centric via AI-Based Methods

This quadrant contains research that employs AI as a methodological tool to construct key variables or generate predictions that answer fundamental accounting questions. Examples include using a transformer-based model to measure disclosure characteristics (e.g., sentiment) and their effects on market outcomes, applying machine learning algorithms to assess fraud risk or predict financial distress, and using deep learning to detect anomalous patterns in financial statements that predict restatements or SEC enforcement actions.

For instance, Jun et al. (2022) use machine learning to validate whether financial statement variables predict stock returns, addressing the value-relevance of fundamental analysis. Drake et al. (2024) employ machine learning to classify analyst forecast revision patterns and examine how forecast diversity affects consensus accuracy and information asymmetry. In these studies, AI serves as a powerful analytical instrument, but the research questions and contributions remain firmly rooted in accounting theory and practice.

### 3.1.2 Accounting-Centric via Traditional Methods

Research in this quadrant investigates accounting outcomes using traditional research methods, while AI provides the environmental context or source of exogenous variation. This includes studies examining the effects and consequences of AI adoption on financial reporting quality, audit quality, and audit fees, as well as natural experiments arising from AI introductions, bans, or service outages.

For example, Fedyk et al. (2022) examine how audit firms' AI workforce investments affect audit quality, fees, and auditor employment using traditional empirical methods, with AI adoption providing the source of variation. Bertomeu et al. (2025) exploit Italy's unexpected ChatGPT ban as a natural experiment to study how AI availability affects analyst forecasting behavior, forecast accuracy, and market information efficiency. In these studies, AI provides the setting or source of variation rather than serving as the analytical tool, with researchers employing conventional econometric approaches to assess its impact on accounting outcomes.

### 3.1.3 AI Centric via AI-Based Methods

This quadrant encompasses research that makes methodological or technical contributions to AI within accounting contexts. These studies advance AI technologies specifically for accounting applications, such as benchmarking anomaly detection algorithms for fraud identification, fine-tuning language models to parse complex financial disclosures, developing specialized architectures for detecting financial statement fraud, or designing multi-modal models that integrate numerical and textual financial data.

For example, Zhang et al. (2022) introduce Explainable AI (XAI) techniques to auditing contexts, demonstrating how these methods enhance the interpretability and transparency of AI models used in audit risk assessment. Huang et al. (2023) develop FinBERT, a domain-adapted large language model fine-tuned on financial texts that substantially outperforms general-purpose models and traditional methods in financial sentiment classification and topic identification. In these studies, the primary contribution lies in advancing AI methodologies and architectures tailored to accounting and finance applications.

### 3.1.4 AI-Centric via Traditional Methods

Research in this quadrant analyzes AI systems and their implications using non-AI methodological approaches. In contrast to Section 3.1.2, which include studies that examine the effects of AI adoption on accounting outcomes, this quadrant focuses on understanding AI adoption itself, e.g., its drivers, barriers, governance challenges, and user perceptions. This includes survey and experimental studies examining AI ethics, governance, and adoption barriers in accounting contexts; conceptual frameworks for understanding AI integration; systematic literature reviews of AI applications in accounting; and research on AI in accounting education.

For example, Kokina et al. (2025) conduct qualitative interviews with 22 audit professionals to examine AI adoption challenges, identifying barriers related to transparency, explainability, bias, and the need for governance guidance. Estep et al. (2024) employ a multi-method approach combining surveys and experiments to investigate how financial executives perceive and respond to AI use by both companies and auditors, revealing that managers' AI adoption decisions influence their receptiveness to auditor AI outputs. In these studies, AI represents the phenomenon under investigation, with researchers using traditional empirical methods to understand adoption patterns, user attitudes, and implementation challenges rather than developing or applying AI technologies themselves.

The four quadrants differ substantially in their technical demands. "AI-Centric via AI-Based Methods" requires the highest level of technical expertise, involving extensive coding and a deep understanding of model architectures. Researchers in this quadrant often train models from scratch or fine-tune pre-trained models for specialized tasks. "Accounting-Centric via AI-Based Methods" demands moderate technical knowledge, as researchers typically apply existing models rather than developing new ones, although they must understand model selection and implementation. "Accounting-Centric via Traditional Methods" requires general familiarity with AI capabilities and limitations to understand its impact on information processing and decision-making. "AI-Centric via Traditional Methods" necessitates conceptual understanding of AI mechanisms and their interaction with organizational dynamics and broader economic impacts.

**Table 1 AI-Accounting Research Framework**

| Research Focus →<br>Methodology ↓ | **Accounting-Centric**<br>(Accounting as research object) | **AI-Centric**<br>(AI as research object) |
|---|---|---|
| **AI-Based Methods**<br><br>(ML/DL/LLM central to analysis) | **Accounting via AI-Based Methods**<br><br>Uses AI techniques to address core accounting questions.<br><br>*Examples*: Transformers to measure disclosure tone and study market effects; ML to detect fraud or predict distress; anomaly detection for restatements. | **AI via AI-Based Methods**<br><br>Advances AI techniques tailored to accounting applications.<br><br>*Examples*: Benchmarking fraud-detection algorithms; fine-tuning LLMs for analyzing financial disclosures; developing multi-modal architectures combining text & numbers. |
| **Traditional Methods**<br><br>(Regressions/ experiments/surveys/interviews /meta studies) | **Accounting via Traditional Methods**<br><br>Examines accounting outcomes using standard empirical or experimental methods, with AI as the environmental context.<br><br>*Examples*: Effects of AI adoption on audit fees or reporting quality; natural experiments from ChatGPT outages or regulatory changes. | **AI via Traditional Methods**<br><br>Analyzes AI systems in accounting using non-AI methods.<br><br>*Examples*: Surveys on AI ethics/governance; experiments on AI-assisted decision-making; interviews on AI adoption barriers; conceptual frameworks; literature reviews on advancements in AI research; studies on AI in accounting education. |

## 3.2 Classification Procedures

To facilitate the reliable application of the framework, we establish manual classification procedures that operationalize our two-dimensional taxonomy. These procedures provide decision rules for determining research centricity and methodological approach, addressing common ambiguities that arise when categorizing interdisciplinary AI-accounting research.

The classification process begins with determining research centricity by examining whether the contribution is framed primarily toward AI methods and systems or toward accounting outcomes. This assessment is validated by identifying the main outcome variables, where model performance metrics indicate AI-centricity, while accounting outcomes suggest accounting-centricity. Next, the primary method is identified by determining whether AI performs the key estimation and measurement tasks or whether inference relies on traditional approaches.

For mixed-method studies that employ both AI tools and traditional inference, classification depends on the primary analytical approach. If AI is used to generate the main construct or estimator for answering an accounting question, the study is primarily classified in the "Accounting via AI" quadrant. For papers making genuine dual contributions, we apply a hierarchical classification approach: first, we prioritize the authors' explicit contribution claims; second, we examine which outcomes serve as dependent variables; third, we assess which contribution would remain novel if the other were removed; and finally, we consider the relative space devoted to each component in the results section. Studies using a design science approach are classified based on whether the artifact is generated primarily through traditional methods or AI-based methods.

Studies examining AI as a setting through introductions, bans, outages, or policy shocks typically default to "Accounting via Traditional methods," as they often feature accounting outcomes as dependent variables and employ traditional methods. Classification shifts to "AI via Traditional methods" only when the dependent variables or contribution focus on the AI system itself, such as trust in AI or model error patterns studied through traditional methods. Simple dictionary-based text analysis is considered a traditional method unless it is benchmarked against modern machine learning or NLP techniques such as embeddings, topic models, or transformers. AI must play a substantive role beyond data preparation to warrant an AI-based method classification.

### 3.3 Demonstrating the Framework: Special Issue Paper Analysis

The seven papers in the recent IJAIS special issue on AI and accounting provide a focused lens through which to demonstrate the utility of our framework. Their distribution across the quadrants highlights the diversity of current approaches and the gaps that remain in AI-accounting research, as shown in Table 2.

**Table 2 Classification of IJAIS Special Issue Papers**

| Paper | Primary Contribution | Classification |
|---|---|---|
| **Veganzones & Séverin** (2025) | Machine learning identification of distinct earnings management strategies through financial profile analysis | Accounting-Centric via AI-Based Methods |
| **Hesse & Loy** (2025) | Sentence-level BERT analysis of MD&A disclosures for bankruptcy prediction with enhanced interpretability | AI-Centric via AI-Based Methods |
| **Levich & Knust** (2025) | Systematic comparison of discriminative vs. generative models for multilingual accounting information extraction | AI-Centric via AI-Based Methods |
| **Lin et al.** (2025) | Visual transformation of financial ratios into images analyzed through CNNs for bankruptcy prediction | AI-Centric via AI-Based Methods |
| **Blondeel et al.** (2025) | Validation of ChatGPT as a secondary coder in qualitative accounting research with implementation protocols | AI-Centric via Traditional Methods |
| **Dong et al.** (2024) | Comprehensive scoping review mapping ChatGPT applications in accounting and finance across three research streams | AI-Centric via Traditional Methods |
| **Fotoh & Mugwira** (2025) | Blind assessment study revealing LLM capabilities and limitations in audit procedures through audit partner evaluations | AI-Centric via Traditional Methods |

- **Accounting via AI (1 paper):** *Veganzones & Séverin* apply machine learning techniques to earnings management, exemplifying how AI methods can enrich traditional accounting inquiries. This contribution underscores the quadrant's potential to leverage AI for refining and testing core accounting constructs.

- **AI via AI (3 papers):** *Hesse & Loy, Levich & Knust, and Lin et al.* represent methodological innovations at the intersection of AI and accounting. Collectively, these studies advance the frontier of LLMs and explainable AI by benchmarking, fine-tuning, and developing hybrid models tailored to financial data. Their prominence in the issue suggests strong momentum toward AI-centric contributions using AI-based methods.

- **AI via Traditional (3 papers):** *Blondeel et al., Dong et al., and Fotoh & Mugwira* illustrate the role of traditional methods, such as surveys, qualitative coding, and conceptual analysis, in evaluating AI systems and their implications for accounting practice. These studies examine the adoption, validation, and ethical dimensions of LLMs, highlighting the value of non-technical inquiry in a domain often dominated by computational approaches.

- **Accounting via Traditional (0 papers):** Notably absent are papers in this quadrant, which would focus on addressing traditional accounting research questions by using AI to provide the context or source of variation. This absence may reflect the authors' perception that AI should play a more substantial methodological or substantive role in their papers to be considered appropriate for the special issue. The lack of such studies could also indicate a gap in the literature and signals a possible frontier for future research.

Taken together, the composition of the special issue illustrates two key patterns. First, there is strong momentum toward AI-centric research, particularly leveraging AI-based methods for methodological advancement. Second, scholars are actively engaging in evaluating LLMs with traditional methods, reflecting the field's recognition of ethical, adoption, and validation challenges. The absence of "Accounting via Traditional" contributions is consistent with both the special issue's perceived focus and the broader diagonal shift observed in AIS journals (to be discussed in the next section), where researchers increasingly move away from traditional designs toward AI-enabled or AI-centric approaches.

### 3.4   Framework Application: Analyzing Recent AI-Accounting Research

We apply our framework to analyze the current state of AI-accounting research, examining how existing studies distribute across the four quadrants. This empirical mapping serves two purposes. First, it validates the framework's comprehensiveness by demonstrating how research fits meaningfully within our categories. Second, it reveals the field's current research patterns and gaps, providing the baseline understanding necessary for identifying collaborative advantages and positioning opportunities that we address in subsequent sections.

To identify papers at the intersection of AI and accounting, we employ a two-step search strategy. We choose to begin our search period in 2022 to capture research that leverages the recent transformative advancements in AI technology, particularly the emergence of LLMs and sophisticated deep learning architectures that have fundamentally changed the landscape of AI applications in business. Our objective is not to provide an exhaustive historical review of AI in accounting since the inception of modern AI, but rather to focus on contemporary research that reflects the current state and near-term trajectory of AI capabilities and their practical applications in accounting. Our focus on peer-reviewed literature ensures methodological rigor and reproducibility at the expense of currency, meaning that emerging developments available only as working papers are not included in our classification.

First, we search the Web of Science Core Collection to capture AI-related research in business disciplines. Our search requires that either the title or abstract include at least one of the following AI-related keywords: "artificial intelligence", "AI", "machine learning", "deep learning", "neural network*", "language model*", or "ChatGPT". We restrict our search to English-language articles indexed in SSCI, ESCI, or SCI-Expanded, and limit the Web of Science categories to Business, Business Finance, or Management. The search covers publications from 2022 through 2025 and was executed on August 7, 2025. To focus specifically on accounting research, we retain only papers published in major accounting journals as identified by the ABDC journal quality list, yielding an initial sample of 381 papers.

Second, from these papers, we retain 101 (about 26.5%), which were published in the following journals:

- AIS journals: *International Journal of Information Systems* (IJAIS),[2] *Journal of Information Systems* (JIS), and *Journal of Emerging Technologies in Accounting* (JETA).
- Non-AIS journals: *Accounting, Organizations and Society* (AOS), *Contemporary Accounting Research* (CAR), *Journal of Accounting & Economics* (JAE), *Journal of Accounting Research* (JAR), *Review of Accounting Studies* (RAST), and *The Accounting Review* (TAR).

We deliberately restrict our analysis to these accounting journals for two key reasons. First, given that our methodology requires manual classification and detailed analysis of each paper to accurately categorize the AI techniques employed and their specific accounting applications, we needed to maintain a manageable scope for the review.[3] Second, by focusing on these journals, we ensure that our analysis captures high-quality, peer-reviewed research that has undergone rigorous evaluation and represents the most influential contributions to the intersection of AI and accounting.

We manually screen and classify each paper according to our framework, based on titles and abstracts, and, in most cases, the full text. We first filter out papers that do not involve AI in a meaningful way, such as those merely mentioning AI when proposing future research opportunities, removing 12 papers. This yields a final sample of 89 papers. Our analysis of these 89 papers shows that AIS journals account for 60 papers (67% of the sample), while non-AIS journals contribute 29 papers (33%). For these papers, each of the authors classified them independently. The inter-rater agreement is 71% (i.e., 63 out of the 89 papers). Differences in classification were resolved through a joint review of the paper. Table 3 shows the distribution of these papers across all journals.[4]

---

[2] It is worth noting that IJAIS has established a strong tradition of publishing innovative research on AI and accounting (Elnakeeb and Elawadly, 2025; Kumar et al., 2020; Sutton et al., 2016). Building on this foundation, JIAIS continues to lead in advancing research in this rapidly evolving field, including publishing the special issue on AI and accounting, and a special issue on advanced technologies in auditing (Black and Gerard, 2025). The latter special issue also featured several papers incorporating artificial intelligence (Kokina et al., 2025; Lee and Tahmoush, 2025; Li and Goel, 2025; Torroba et al., 2025). Beyond its leadership in AI research, IJAIS's overall impact has risen substantially, establishing it among the highest-impact outlets in accounting research (Grabski and Leech, 2023).

[3] Even though state-of-the-art LLMs possess the capability to classify research papers according to predefined rubrics, most publishers prohibit the use of their published content with LLMs due to copyright and licensing restrictions.

**Table 3 Distribution of Papers Across All Journals**

|  | Accounting-Centric | AI-Centric | Total |
|---|---|---|---|
| *AI-Based Methods* | 28 (31.5%) | 15 (16.9%) | *43 (48%)* |
| *Traditional Methods* | 3 (3.4%) | 43 (48.3%) | *46 (52%)* |
| *Total* | *31 (35%)* | *58 (65%)* | *89 (100%)* |

While there is a relatively even distribution across methodology (48% AI-based method vs. 52% for traditional methods), there is a stronger focus on AI-centric papers (35% for Accounting-centric vs. 65% for AI-centric). Additionally, we observe a sharp contrast along the diagonal, with a notable concentration of papers in the "AI via AI" cell (16.9%) compared to the "Accounting via Traditional" cell (3.4%). This diagonal imbalance reveals differing technical skill requirements across research approaches. Traditional accounting methods rely on established quantitative skills, while AI-centric research demands competencies in machine learning, programming, and newer computational methods. This pattern signals a fundamental transformation in how accounting research is conducted and the expertise needed to contribute to the field.

Tables 4 and 5 capture the distribution of these papers from AIS and non-AIS journals, respectively. The distribution patterns reflect the distinct research priorities characteristic of AIS and non-AIS journals. Each group exhibits different concentration patterns across the framework's four quadrants.[5, 6] AIS journals exhibit a relatively strong preference for AI-focused papers (80%), with the highest concentration (58%) in the lower-right quadrant (AI-Centric via Traditional Methods). This reflects their emphasis on critically evaluating AI systems through experiments or understanding their adoption patterns through surveys and field studies.

---

[4] We acknowledge that this distribution may be influenced by our selection of journals. While we include all major AIS-focused journals (IJAIS, JIS, and JETA), we include only six general accounting journals (AOS, CAR, JAE, JAR, RAST, and TAR). This asymmetric representation may bias our distribution in several ways. First, since we capture the full universe of major AIS journals but only a subset of general accounting journals, we likely overrepresent AI-centric research relative to what would be observed if all general accounting journals were included. Second, the AIS journals, by their specialized nature, are predisposed to publish research at the intersection of technology and accounting, particularly AI applications, which may inflate the proportion of papers in the AI-centric categories. Conversely, our exclusion of other general accounting journals that might contain AI-related research could understate the true extent of AI adoption in mainstream accounting research. Future research could either focus exclusively on general accounting journals or include a more comprehensive set of journals to validate these distributional patterns.

**Table 4 Distribution of Papers from AIS Journals**

|  | Accounting-Centric | AI-Centric | Total |
|---|---|---|---|
| *AI-Based Methods* | 12 (20%) | 13 (22%) | *25 (42%)* |
| *Traditional Methods* | 0 (0%) | 35 (58%) | *35 (58%)* |
| *Total* | *12 (20%)* | *48 (80%)* | *60 (100%)* |

There is a relatively even split among the papers that use AI-based methods between accounting-centric (20%) and AI-centric (22%). The second-highest concentration (22%) is in the upper-right quadrant (AI via AI). This distribution underscores the dual role of AIS journals, as sophisticated adopters of AI for accounting applications and advancing AI technologies.

**Table 5 Distribution of Papers from Non-AIS Journals**

|  | Accounting Centric | AI Centric | Total |
|---|---|---|---|
| *AI-Based Methods* | 16 (55%) | 2 (7%) | *18 (62%)* |
| *Traditional Methods* | 3 (10%) | 8 (28%) | *11 (38%)* |
| *Total* | *19 (65%)* | *10 (35%)* | *29 (100%)* |

In contrast, non-AIS journals exhibit a relatively strong preference for accounting-centric papers (65%), with 55% of the papers in the upper-left quadrant (Accounting via AI). This dominance reflects a strong focus on leveraging AI as a methodological tool for traditional accounting inquiries, such as using AI-based NLP for disclosure analysis, machine learning for fraud detection, and deep learning for audit research. Their limited presence in AI-centric quadrants (only 35% combined in the right column) suggests that AI is viewed primarily as an advanced analytical instrument rather than a subject of inquiry.

---

[5] It is important to note that while these distributional differences are observed, they reflect authors' self-selection rather than editorial preferences. IJAIS and other AIS journals welcome and actively publish research across all four quadrants, as do the general interest journals. Authors naturally gravitate toward outlets whose readership aligns with their research contributions, resulting in these observed patterns.

[6] We acknowledge that publication timelines vary across journals, meaning that papers appearing in the same year across AIS and non-AIS outlets may represent research initiated at substantially different times. This temporal variation may influence the observed distributional patterns across journal types.

An interesting distribution pattern emerges when contrasting the diagonal cells between the two sets of journals. While differences in publication timelines across outlets may partially influence these patterns, the contrast remains notable. AIS journals display an ascending diagonal distribution (from 0% in the lower-left to 22% in the upper-right), while non-AIS journals show a descending diagonal pattern (from 10% to 7%) across the same cells. These diagonal patterns reveal fundamentally different approaches: AIS journals embrace AI as part of their evolution, often developing or refining AI technologies tailored to accounting. Non-AIS journals, meanwhile, treat AI as an advanced tool to enhance traditional inquiries, maintaining a clear accounting-centric identity.

## 4. Research Opportunities: Leveraging Collaborative Advantage

### 4.1 Collaborative Advantage in the AI-Accounting Research Landscape

Our framework categorizes AI-accounting research into four quadrants. Research opportunities exist in each quadrant, and identifying these opportunities requires understanding one's collaborative advantages. Alles et al. (2008) and Alles & Gray (2025) emphasize comparative advantage as a strategic positioning tool for researchers. We reframe this as collaborative advantage and recognize that the potential research partnerships now extend beyond the traditional academic-industry duality. The landscape now includes at least three groups: (1) accounting researchers studying AI and accounting; (2) industry researchers developing and deploying AI; (3) computer science researchers developing general AI tools, which may be adapted or applied to the accounting context. AI agents themselves represent a special case. They can now perform many research-related tasks but are best understood as advanced tools rather than collaborators. We discuss AI agentss in the next section, which focuses on how advances in AI are reshaping the research process.

Each of the three groups brings unique strengths and faces distinct limitations, as summarized in Table 6. Accounting researchers possess deep domain expertise, institutional knowledge, and theoretical grounding, but face slower publication cycles and limited computational resources. Industry researchers have access to proprietary data, rapid deployment capabilities, and real-world problem contexts, but are constrained by client needs and competitive pressures that limit transparency. Computer science researchers operate at the methodological frontier with state-of-the-art techniques and faster iteration cycles, but lack the accounting domain knowledge and the motivation to customize the technology for accounting applications.

**Table 6 Comparative Strengths Across Research Communities**

| Task | Accounting Researchers | Industry Researchers | CS Researchers |
| --- | --- | --- | --- |
| Accounting domain depth | Deep institutional & regulatory knowledge | Practice-specific expertise; compliance focus | Surface-level; requires collaboration for nuance |
| Idea generation & theorizing | Deep domain priors; theory-driven; institutional knowledge | Practice-proximate; client constraints; real-world problem focus | Novel methods; algorithmic innovation; less domain-specific |
| Data access & legitimacy | Academic partnerships; field site access; enhanced legitimacy via institutional IRB approval | Proprietary systems; production data; large-scale deployment data | Public corpora; benchmark datasets; simulated data |
| Method innovation | Established econometric methods; adapting ML cautiously | Rapid deployment; applied analytics; large-scale field testing | Frontier algorithmic development; state-of-the-art techniques |
| Reproducibility & governance | Traditional peer review; working papers; replication standards | Corporate IP protection; limited transparency | Open-source culture (selective); preprints; reproducible benchmarks |
| Speed & iterations | Slow (publication cycles: 2-5 years) | Fast (quarterly/annual product cycles) | Fast (conference cycles: 6-12 months; rapid prototyping) |
| Theoretical contribution | High; causal identification emphasized; mechanism exploration; result robustness | Low; proof-of-concept; performance optimization | Medium; architectural innovations; interpretability; model transferability |

## 4.2    Research Opportunities

Using our 2×2 framework (Table 1), we identify four strategic positions where accounting researchers can leverage their collaborative advantages relative to industry and computer science researchers. Below, we propose research opportunities tailored to each quadrant. While some questions may echo prior studies, the rapid advancement of GenAI, with its unprecedented capabilities in natural language understanding, content generation, and complex reasoning, necessitates a fresh examination of both classic and emerging issues in accounting context.

### 4.2.1    AI-Centric via AI-Based Methods

While academics cannot compete with industry's computational resources for developing foundational AI models, they possess collaborative advantages in creating specialized AI applications for accounting contexts. Unlike industry solutions designed for profit maximization and proprietary advantage, academic researchers can develop or refine open-source algorithms for tasks such as audit sampling, risk assessment, or anomaly detection that prioritize transparency, interpretability, and compliance with professional standards.

Accounting researchers can also conduct rigorous comparative studies of different AI architectures for accounting tasks. This is a type of research that vendors tend to avoid, as it might reveal their solutions' limitations. Where industry researchers optimize for deployment speed and CS researchers focus on algorithmic novelty, accounting researchers can contribute specialized techniques that balance technical sophistication with domain-specific requirements such as auditability, interpretability, and regulatory compliance. This quadrant enables researchers to advance AI techniques specifically tailored to accounting applications, leveraging their understanding of institutional constraints that neither the industry nor CS researchers fully appreciate.

These advantages enable distinctive research contributions. For example, accounting researchers can develop explainable AI systems for going-concern predictions that auditors can defend to stakeholders, create multimodal models integrating financial and non-financial data with complete audit trails, and fine-tune LLMs for disclosure analysis using domain-specific training that reflects accounting conventions and regulatory requirements.

### 4.2.2 AI-Centric via Traditional Methods

This quadrant represents perhaps the strongest advantage for accounting academics: the independent, critical evaluation of AI technologies using established research methods. Industry has little incentive to fund surveys revealing accountant resistance to AI tools, experiments demonstrating when AI recommendations lead to poor judgments, or field studies documenting implementation failures. Even if they do so, they have no incentive to publicize their findings. Computer science researchers, focused on algorithmic performance metrics, are less likely to examine the organizational, professional, and societal implications of AI adoption.

Accounting researchers' distinctive advantages, such as IRB capacity for human subject research and independence from commercial interests, enable several categories of critical inquiry. These advantages remain particularly valuable precisely because the field develops rapidly. Independent research can provide unbiased assessment of societal impacts and ethical implications that commercial development may not adequately address. Key examples may include:

- Studies that examine practitioner perceptions, resistance factors, and organizational barriers to AI implementation across audit, financial reporting, and management accounting contexts;
- Behavioral and judgment research investigating how AI recommendations affect professional skepticism, judgment quality, over-reliance patterns, user effort and engagement, and decision-making under uncertainty—questions that require controlled experiments with domain experts;

- Implementation and failure analysis using field studies, case studies, and interviews to document why AI tools succeed or fail in specific organizational and professional contexts, revealing boundary conditions that vendors rarely publicize;
- Professional and ethical implications research addressing AI's effects on skills development, professional identity, algorithmic bias in accounting applications, and governance mechanisms—issues that require both technical understanding and deep institutional knowledge;
- Performance validation studies that rigorously test vendor claims against actual outcomes using archival data and field evidence, providing the independent evaluation that commercial interests discourage.

While prior research has addressed some of these questions for earlier AI technologies, many remain unanswered for recent advances such as GenAI, which introduce fundamentally different capabilities and risks. These research directions leverage accounting researchers' unique position to provide an essential counterweight to industry marketing claims and fill gaps that CS researchers overlook, identifying risks and boundary conditions that neither technology vendors nor computer science academics have the incentive or contextual expertise to examine.

### 4.2.3 Accounting-Centric via AI-Based Methods

This quadrant exemplifies the unique position of academics at the intersection of methodological innovation and theoretical grounding. Here, accounting researchers leverage their deep domain knowledge while applying AI methods to longstanding accounting questions. Unlike industry data scientists who may lack accounting expertise or practitioners constrained by client confidentiality, academics can use machine learning to test fundamental theories about earnings management, disclosure quality, or audit effectiveness, among others.

Accounting researchers' collaborative advantages in this quadrant stem from their ability to combine AI techniques with theoretical frameworks and causal inference methods. While CS researchers might develop superior technical methods and industry researchers have better data access, accounting researchers possess the expertise necessary to formulate theoretically-grounded research questions, design appropriate identification strategies, and interpret findings within established theoretical frameworks. Research directions that leverage this combination include:

- AI-enhanced measurement of accounting constructs: Natural language processing and transformer models can generate novel measures of disclosure quality, audit report features, management tone, or governance characteristics that were previously difficult to quantify systematically. These measures enable researchers to test theories using traditional econometric methods with richer data than manual coding allows.
- Machine learning for pattern detection and prediction: AI methods can identify nonlinear relationships, detect anomalies, or generate predictions (such as financial distress, earnings management likelihood, or audit risk) that serve as inputs to causal inference designs. The predictive capability complements, but does not replace, the need for rigorous identification strategies to test competing theories.
- Scalable analysis of textual and unstructured data: AI enables researchers to analyze decades of financial disclosures, audit reports, earnings call transcripts, or regulatory filings at scale, creating opportunities to test theories about disclosure strategies, information processing, or market reactions across broader samples and longer time periods than manual analysis permits.
- Decomposition and residual analysis: Machine learning models can decompose variation in accounting outcomes (such as audit fees, cost behavior, or accruals) to isolate unexplained components, which researchers can then study using traditional methods to test theories about underlying economic mechanisms.
- Cross-domain integration: AI methods facilitate combining financial data with non-financial information (such as news sentiment, social media, supplier networks, or environmental metrics) to test theories about information spillovers, stakeholder influences, or real effects of disclosure, requiring domain knowledge to ensure appropriate measurement and interpretation.

This combination of AI techniques with accounting theory and traditional research designs addresses core questions that neither industry (focused on applied problems) nor CS researchers (lacking domain depth) would naturally pursue. The value lies not in AI methods alone, but in accounting researchers' ability to deploy these tools within rigorous research designs that enable strong inferences and connect empirical findings to theoretical predictions.

### 4.2.4 Accounting-Centric via Traditional Methods

This quadrant represents a relatively narrow research space where AI provides environmental variation or contextual background for studying traditional accounting questions. Here, AI is involved superficially, not as a methodological tool, not as the primary object of study, and not requiring deep AI expertise. Instead, AI-related events, regulations, or exogenous shocks create natural experiments or provide variation for examining core accounting phenomena using established empirical methods. It is important to note that research examining how AI fundamentally changes workplace dynamics, such as effects on engagement, team interactions with AI agents, accountability structures, job security, or psychological safety, would typically belong in other quadrants (particularly behavioral and judgment research in "AI via Traditional") because AI is the primary phenomenon being studied, not merely contextual variation. The key distinguishing feature of this quadrant is that AI provides variation, timing, or context for studying accounting phenomena, but the research questions, methods, and findings are fundamentally about accounting, not about AI technologies, AI adoption processes, or AI-enabled measurement.

While potentially the smallest quadrant in terms of pure methodological scope, industry researchers have no interest in these theoretical questions, and CS researchers lack both the motivation and the accounting expertise to formulate or pursue them. The advantage of accounting academics lies in recognizing how AI-related events can serve as instruments or sources of variation for studying fundamental accounting relationships.

Researchers can exploit AI-related regulations as exogenous shocks, such as examining how AI disclosure requirements affect information asymmetry, studying how university ChatGPT bans impact accounting student learning outcomes, or analyzing how geographic variation in AI regulations or accessibility affects traditional accounting and auditing outcomes.

Academics can use AI events as instrumental variables, such as leveraging ChatGPT's release date as a discontinuity for studying changes in analyst report quality, exploiting AI vendor outages to study their impact on information processing and market outcomes, or using variation in firms' AI exposure to examine differences in earnings quality or management forecast accuracy.

Over the longer term, researchers can examine institutional responses to AI, such as analyzing how standard-setters modify financial reporting requirements as AI becomes prevalent in business models, how audit firms restructure audit teams in response to AI adoption by themselves or their clients, or how professional licensing requirements evolve as AI capabilities become widespread.

## 4.3    Strategic Implications

Within this evolving landscape, accounting researchers enjoy multiple avenues for strategic positioning. By situating their contributions within one of the four quadrants, they can advance accounting knowledge through AI-based methods, evaluate the implications of AI adoption with traditional approaches, or push the boundaries of AI itself in accounting contexts. This flexibility creates a collaborative advantage for scholars who combine deep domain expertise with fluency in AI tools, allowing them to bridge the gap between accounting-centric inquiries and AI-centric innovations.

However, strategic positioning requires understanding the broader research landscape. Computer science researchers and industry laboratories are producing technically advanced models at a pace that demands accounting researchers identify their contributions. Industry researchers' access to proprietary data and rapid deployment cycles enables them to address applied problems faster than academic publication timelines allow. Accounting researchers' contribution lies in leveraging what they uniquely possess: deep institutional and regulatory knowledge, established theoretical frameworks, causal inference expertise, independent evaluation capabilities, and legitimacy within the accounting profession. By strategically positioning their research within the framework's quadrants, accounting scholars can identify high-value opportunities where strategic advantages through collaboration create the most value.

While exploring these opportunities, researchers should be aware of one important development that is reshaping the research landscape: the emergence of GenAI as a research assistant. This technological shift not only changes what we study but fundamentally transforms how we conduct research itself. We explore this adaptation in the next section.

## 5. Adapting to AI-Augmented Research

### 5.1 Comparative Advantages of Human Researchers vs AI Agents

So far, we have discussed research opportunities presented by AI. Yet beyond serving as a research subject or methodological tool, recent developments in GenAI, particularly LLMs like ChatGPT, Claude, and domain-specific variants, are fundamentally transforming how researchers conduct research itself (Barrick et al., 2025; de Kok, 2025; Vakilzadeh and Wood, 2025; Wang and Wang, 2024). Although we do not expect fully autonomous AI agents capable of conducting research with minimal human supervision in the near term, GenAI models as research assistants are already reshaping the research workflow across multiple stages. Almost two-thirds of accounting researchers surveyed already use these tools in their research (Barrick et al., 2025), for example, 53% use GenAI for rewriting or editing drafts, 31% for conducting literature reviews, and 28% for generating code for data analysis.

These models possess capabilities that span the entire research pipeline: generating research ideas through novel recombination of existing questions, conducting systematic literature reviews, designing research protocols, processing and analyzing data, creating visualizations, interpreting results, and drafting manuscripts (Barrick et al., 2025). However, their current capabilities vary significantly across tasks (Dowling and Lucey, 2023; Keloharju and Keloharju, 2025). Understanding where AI models or agents complement versus substitute for human researchers is critical for leveraging these tools effectively.

Table 7 compares the capabilities of AI agents with human researchers across key research tasks. The table is organized into two panels that reflect the natural flow of the research process. Panel A examines upstream tasks, i.e., the conceptual and design phase, where researchers frame questions, synthesize literature, and establish methodologies. Panel B addresses downstream tasks, i.e., the execution, analysis, and dissemination phase, where researchers implement their designs, analyze data, interpret results, and communicate findings.

**Table 7 Comparing AI Agents and Human Researchers in the Research Process**
**Panel A: Upstream (Framing and Design)**

| Task | Human Researchers | AI Agents | Optimal Division of Labor |
|---|---|---|---|
| Domain expertise & tacit knowledge | Years of accumulated experience; institutional knowledge; professional judgment | Training data cutoff; no lived experience; cannot acquire new tacit knowledge through practice | Humans are for deep expertise; AI is useful for cross-domain knowledge retrieval |
| Idea generation & theorizing | Deep contextual understanding; ground-breaking insights; slow deliberation; domain expertise | Rapid recombination of existing concepts; identifying gaps in literature; risk of conventional/derivative ideas; lacking true creativity | AI for breadth and initial exploration; humans for theoretical depth and novel frameworks |
| Literature review & synthesis | Selective reading; potential of confirmation bias; time-intensive; deep comprehension | Comprehensive coverage; fast summarization; pattern detection across large corpora; may miss nuance or misinterpret context | AI for comprehensive scanning and organization; humans for critical evaluation and theoretical integration |
| Research design & methodology | Domain-specific expertise; causal inference knowledge; contextual judgment | Can suggest established designs; retrieves appropriate methods from training data; limited adaptation to novel contexts | AI for standard designs and method retrieval; humans for custom designs requiring domain judgment |
| Ethical oversight | IRB training; professional standards; contextual judgment | No independent ethical judgment; can flag potential issues if prompted; requires human governance | Humans are essential for all ethical decisions; AI can assist with checklist compliance |

**Table 7 Comparing AI Agents and Human Researchers in the Research Process**

**Panel B: Downstream (Execution, Analysis & Dissemination)**

| Task | Human Researchers | AI Agents | Optimal Division of Labor |
|---|---|---|---|
| Pre-processing & cleaning of unstructured data | Manual coding; labor-intensive; prone to fatigue; good at handling context-dependent ambiguities | Rapid initial processing; can quickly identify patterns and structure; may miss nuanced contextual cues without clear guidelines | AI for initial processing and pattern detection; humans for ambiguous cases and validation |
| Statistical analysis & coding | Methodological expertise for theory-driven model specification; appropriate test selection based on research design; interprets assumption violations contextually; understands when violations matter substantively vs. can be ignored; slower code execution | Executes code rapidly; retrieves diverse statistical methods; can suggest alternative approaches; may apply methods mechanically without verifying appropriateness | Humans specify research questions and theoretical models; AI generates code and suggests alternative methods (broadening methodological toolkit); AI runs diagnostics; humans interpret diagnostic results; humans make final decisions on model selection and interpretation |
| Data visualization | Design judgment; audience awareness; theoretical framing | Rapid generation of multiple visualization options; follows standard conventions; may lack audience-specific customization | AI for initial drafts and iteration; humans for final design decisions aligned with narrative |
| Results interpretation | Theoretical grounding; causal reasoning; connects findings to literature | Pattern description; statistical interpretation; may over-interpret correlations | Humans are essential for theoretical interpretation; AI for descriptive summaries and alternative explanations |
| Manuscript drafting | Narrative coherence; field-specific conventions; voice and style | Fast drafts; clear structure; generic academic prose; may lack argumentative depth or field-specific nuance | AI for structural scaffolding and routine sections; humans for argumentation, framing, and revision |
| Reproducibility & documentation | Variable documentation practices; often incomplete | Can generate detailed code documentation and workflows; requires human oversight to ensure accuracy | AI for automated documentation generation; humans for verification and completeness |
| Speed & iteration | Slow; constrained by cognitive limits and time | Very fast; can explore multiple approaches simultaneously; requires human supervision and validation | AI for rapid prototyping and exploration; humans for strategic direction and quality control |

The comparative advantages reveal a complementary relationship rather than substitution. AI agents excel at tasks that require speed, scale, and pattern recognition across large information spaces, such as literature scanning, data processing, code generation, and routine analysis. Human researchers remain essential for tasks requiring judgment, creativity, causal reasoning, theoretical depth, ethical oversight, and contextual understanding.

Most critically, AI agents operate as tools that amplify human capabilities rather than as autonomous researchers. They require what we term "human scaffolding," i.e., clear instructions, validation checkpoints, and continuous oversight, at least in the near term. An AI agent can process thousands of financial disclosures in minutes, but a human researcher must design the coding scheme, validate the results, and interpret the patterns within theoretical frameworks. An AI agent can draft a literature review, but a human researcher must critically evaluate the synthesis, identify gaps, and position the contribution.

This division of labor has important implications for accounting researchers. First, researchers who effectively leverage AI assistants can dramatically accelerate certain research stages, particularly data processing, literature review, and coding tasks that previously required substantial RA time. This speed advantage may partially offset the resource disparities with industry and CS researchers discussed earlier in terms of research productivity. Second, human advantage increasingly lies in higher-order cognitive tasks: theoretical development, causal inference, contextual judgment, and critical evaluation. Third, new skills become essential: prompt engineering, AI output validation, and knowing when to trust versus verify AI-generated content.

For the AI-accounting research landscape, GenAI democratizes certain research capabilities (e.g., processing unstructured data at scale) while intensifying competition in others (e.g., rapid prototyping). Accounting researchers can leverage these tools to enhance productivity in data-intensive quadrants (accounting-centric via AI-based methods) while maintaining their collaborative advantage in theory-driven interpretation and institutional knowledge. However, as AI capabilities continue advancing, the boundary between human-essential and AI-capable tasks will shift, requiring ongoing strategic adaptation in how researchers establish their contributions.

## 5.2 Implications for Research Teams and Career Development

The transformation of research workflows through AI has profound implications for how a research team is organized and how scholars develop throughout their careers. Understanding these shifts is essential for strategic positioning at both individual and institutional levels.

As AI automates many labor-intensive tasks, such as manual data collection, systematic literature reviews, initial coding, and preliminary drafting, the traditional research team structure is evolving. Projects that once required large teams of doctoral students and research assistants can now be completed by smaller, more agile groups leveraging AI tools. This shift does not eliminate the need for human researchers but rather reallocates human effort toward higher-value activities, such as research design, theoretical development, quality control, and interpretation.

Large research teams will remain essential for multidisciplinary projects requiring diverse expertise, complex field studies involving extensive stakeholder engagement, or initiatives where human judgment and contextual understanding are valuable. However, for many empirical studies, particularly those involving archival data analysis, text processing, or pattern detection, we expect that small teams augmented by AI tools will become the norm rather than the exception.

The integration of AI creates distinct challenges and opportunities at every career stage, requiring scholars to adapt their strategies to maintain a collaborative advantage. Early-career researchers such as junior scholars and PhD students face both the greatest opportunities and the most significant threats. GenAI agents now demonstrate capabilities in literature review, data collection, and preliminary analysis (de Kok, 2025; Vakilzadeh and Wood, 2025; Wang and Wang, 2024), tasks that traditionally formed the core apprenticeship experience for doctoral students.

However, the displacement risk comes not primarily from AI itself but from peers who integrate AI more effectively into their workflows, just as accountants may be displaced not by AI directly but by colleagues who leverage AI (Boritz and Stratopoulos, 2023). As documented by Noy & Zhang (2023) and Brynjolfsson, Li, & Raymond (2025), productivity gains from GenAI are especially pronounced for less-experienced workers, suggesting that junior scholars who master these tools can rapidly advance their research agendas. Consistent with this pattern, Filimonovic et al. (2025) find that in the social and behavioral sciences, GenAI particularly enhances the performance of early-career researchers. Competitive advantage will increasingly depend on the ability to design, supervise, and interpret AI-driven analyses rather than on manual execution of routine tasks.

Junior scholars can develop AI proficiency by publishing AI-assisted extensions of dissertation work or replication studies that benchmark new methods against established findings. Success at this stage requires balancing two priorities: developing deep domain expertise that AI cannot replicate, and mastering AI tools that amplify research productivity.

Mid-career scholars' advantage lies in leveraging established networks, institutional knowledge, and theoretical expertise to shape projects where AI complements rather than substitutes human insight. This might involve leading projects in the lower-right quadrant (AI-centric via traditional methods), where methodological expertise and field access remain crucial, or directing research in the upper-middle quadrant (accounting-centric via AI-based methods) where deep domain knowledge guides AI application.

The central challenge for established scholars is adapting research practices built on traditional methods to AI-augmented workflows. Those who successfully retool will maintain competitiveness by combining accumulated domain expertise with enhanced productivity. Those who resist integration risk being outpaced not because their research questions lack merit, but because their methods cannot match the speed and scale of AI-enhanced approaches.

Experienced researchers occupy a pivotal position in guiding the field's evolution. Their collaborative advantage is less in directly adopting new tools but more in providing intellectual leadership: framing debates on AI's role in accounting research, ensuring investigations remain anchored in theory and practice, and mentoring younger scholars to harness AI responsibly. By shaping research agendas, reinforcing standards of rigor, and synthesizing emerging findings, senior researchers can balance technological innovation with conceptual clarity, rigor, and professional responsibility.

Senior scholars can contribute by identifying research opportunities that require accumulated wisdom, e.g., recognizing practical problems that theory has not addressed, connecting disparate literatures in novel ways, formulating questions that challenge existing paradigms, and identifying emerging phenomena before they become apparent in data. These creative acts remain distinctly human capabilities where experience and deep immersion in the field provide valuable advantages.

## 5.3 Reforming Doctoral Training for the AI Era

### 5.3.1 Opportunities and Threats

The competitive dynamics described above make doctoral training the pivotal lever for shaping accounting research's future. PhD programs are uniquely positioned to equip the next generation with the skills required to thrive in an AI-driven research environment. Because AI capabilities evolve rapidly, our recommendations adopt a short prediction window of four to five years, which matches the typical PhD program duration, providing a forward-looking yet realistic horizon. Attempting longer-term forecasts is impractical given the rapid emergence of new capabilities.

GenAI creates significant opportunities for doctoral education. Students can save substantial time on traditionally labor-intensive tasks such as literature review, manual data collection, hand-coding accounting constructs, and programming, thus accelerating their entry into substantive research questions. This efficiency advantage allows doctoral candidates to pursue more ambitious, data-intensive projects that would have been impractical within traditional PhD timelines. Beyond productivity gains, AI opens new methodological frontiers. Students can now use LLM agents to extract and code information from unstructured disclosures at scale, validate hand-coded samples against AI-generated classifications, automate alternative variable specifications for sensitivity analysis, or process thousands of corporate disclosures to explore patterns infeasible through manual coding. Students may also use AI agents to help design experiments or questionnaire instruments or to pre-test their experimental designs.

However, these opportunities come with significant challenges. As AI commoditizes standard research tasks such as data collection, literature reviews, coding, and initial drafting, the bar rises for higher-order capabilities where humans retain a comparative advantage, including creative idea formation, causal identification, institutional insight, and theoretical depth. Reliability and reproducibility risks from hallucinations, model drift, and prompt sensitivity necessitate careful documentation of model versions, prompts, random seeds, and validation protocols. Finally, asymmetric access to computational resources may widen capability gaps, potentially disadvantaging programs with fewer resources.

### 5.3.2 Curriculum and Pedagogical Reform

Rather than relying on the traditional apprenticeship model, where doctoral students contributed primarily through labor-intensive tasks like data collection, coding, and literature review, PhD programs must pivot toward cultivating higher-order competencies that allow students to supervise, integrate, and critically interpret AI-driven analyses.

A promising pedagogical approach involves replication-based training, where students re-examine established accounting studies using GenAI. This method offers multiple advantages. It builds familiarity with established research, provides hands-on experience with AI tools, reveals practical challenges of integrating AI into established research designs, and develops dual competencies in both traditional and AI-augmented methods.

Training students to develop and maintain specialized AI tools tailored to their research domains is essential for building cumulative advantage throughout their careers. By customizing general-purpose AI systems with domain-specific knowledge, prompts, and workflows, scholars can create personalized assistants that evolve alongside their work, preserving institutional memory and building on insights across projects and career stages.

Just as PhD programs have traditionally relied on economics departments for training in economic theory and econometric methods and on psychology departments for behavioral theory and experimental design, the next generation may need to partner with computer science departments for AI training. The core curriculum should include several components. First, AI fundamentals covering model architectures, training processes, capabilities, and limitations provide essential conceptual grounding. Second, students must develop skills in prompt engineering (i.e., designing effective instructions for AI systems and validating outputs), as well as AI-assisted research methods that integrate AI into data collection, processing, analysis, and interpretation workflows. Third, training should emphasize reproducibility protocols for documenting model versions, prompts, and validation procedures. Finally, the curriculum must address ethical considerations, including understanding bias, fairness, transparency, and professional responsibility in AI-augmented research.

Students' backgrounds will shape their positioning and research domains. Those trained in business or management may gravitate toward accounting-centric research, while those from STEM disciplines may specialize in AI-centric contributions. Programs should accommodate this diversity while ensuring all students develop both domain expertise and AI fluency. Throughout this reformed curriculum, emphasis must remain on capabilities where humans retain a comparative advantage, e.g., connecting disparate literature, identifying emerging phenomena, formulating interesting research questions, and embedding findings within broader theoretical frameworks. While AI amplifies human productivity by processing vast datasets and enabling new methodological possibilities, the intellectual activities, involving the questions asked, the frameworks applied, and the interpretations offered, remain distinctly human contributions.

### 5.3.3 Strategic Priorities for Programs and Researchers

Successfully implementing these reforms requires coordinated action at both the individual and institutional levels. For individual researchers and students, three priorities prove foundational. First, deep domain-specific knowledge, including institutional details, regulatory environments, professional practices, and theoretical foundations, remains paramount. This expertise cannot be automated and forms the basis for formulating meaningful research questions and interpreting findings within the proper context. Cultivating creativity and originality, the ability to identify gaps, challenge assumptions, and connect ideas in novel ways, represents a source of comparative advantage that – at least for the short term - resists automation.

At the institutional level, universities must make strategic investments to support this transition. Providing computational resources, data access, and AI training programs prevents capability gaps from widening due to asymmetric resource availability. Equally important, given the rapid evolution of AI capabilities, is fostering adaptability by emphasizing learning how to learn. This approach equips students with foundations that allow them to adapt as technologies advance rather than training on specific tools likely to become obsolete.

By reforming doctoral education along these lines, the accounting research community can prepare scholars who leverage AI's productivity gains while preserving and enhancing the distinctly human contributions that define valuable research. This balanced approach positions the next generation to thrive in an AI-augmented research environment while maintaining the field's intellectual standards and societal relevance.

## 6. Conclusion

The seven papers in this special issue demonstrate the breadth and value of AI-accounting research, advancing the field through methodological innovation in AI techniques, AI-enhanced accounting inquiry, and critical evaluation of AI systems. This very diversity, while demonstrating the field's richness, raises fundamental questions: Where do these contributions fit relative to each other and the broader literature? What opportunities remain underexplored? How should researchers navigate this increasingly complex landscape?

Addressing these questions requires a systematic framework. This paper proposes a framework for understanding and navigating the rapidly evolving intersection of artificial intelligence and accounting research. By organizing AI-accounting research along two dimensions, i.e., research focus (accounting-centric versus AI-centric) and methodological approach (AI-based methods versus traditional methods), we provide a parsimonious lens through which to classify existing contributions, identify underdeveloped research domains, and position future inquiries strategically.

Applying the framework to the special issue papers reveals their positioning across three quadrants. One paper uses AI methods to enrich traditional accounting inquiries into earnings management ("Accounting via AI"). Three papers advance methodological innovations in AI techniques tailored to accounting contexts ("AI via AI"), while three employ traditional methods to critically evaluate AI systems' adoption, validation, and ethical dimensions ("AI via Traditional"). The composition reveals strong momentum toward AI-centric research and active engagement in evaluating AI systems using traditional methods.

We extend the framework to classify 89 recent AI-accounting papers published in leading AIS and non-AIS journals between 2022 and 2025. Research patterns differ substantially across journal types: AIS journals concentrate on AI technologies as primary subjects of inquiry, emphasizing their evaluation and technical development, whereas non-AIS journals prioritize fundamental accounting questions and employ AI to enhance traditional analyses.

The framework's strategic value extends beyond descriptive classification to prescriptive guidance. By analyzing collaborative advantages relative to industry and computer science researchers, we identify four distinct research domains where accounting scholars can make unique contributions. These range from developing specialized AI algorithms tailored to institutional constraints, to conducting independent critical evaluations of AI systems through experiments and field studies, to testing fundamental theories using AI-enhanced measurement at large scales. Each quadrant offers opportunities for accounting researchers to leverage their distinctive capabilities, such as deep institutional knowledge, theoretical grounding, causal inference expertise, and independent evaluation capacity, in ways that neither industry practitioners nor computer science researchers would naturally pursue.

Yet recognizing opportunities alone proves insufficient without understanding how AI transforms the research process itself. The research landscape increasingly includes AI agents themselves, which now demonstrate capabilities across the entire research pipeline. While AI agents excel at tasks requiring speed, scale, and pattern recognition, human researchers retain essential advantages in judgment, creativity, theoretical depth, and causal reasoning. This division of labor carries profound implications for doctoral training, requiring curriculum reforms that integrate AI fundamentals while cultivating capabilities where humans retain a comparative advantage: connecting disparate literatures, formulating theoretically grounded research questions, and embedding findings within broader frameworks.

Looking forward, the trajectory of AI-accounting research will be shaped by continued advances in AI capabilities, regulatory developments surrounding AI governance, and whether accounting researchers successfully leverage their strategic advantages. The framework presented in this paper provides tools to navigate these dynamics strategically. The scholars who will shape the field's future are those who combine deep domain expertise with AI fluency, recognize where human judgment remains valuable, and strategically position their contributions where collaborative advantages are strongest. The transformation is underway; the question is not whether accounting research will be affected by AI, but rather how strategically we navigate this transformation to ensure the field's continued vitality and relevance.

**Declaration of generative AI and AI-assisted technologies in the writing process**

While preparing this manuscript, the authors used ChatGPT and Claude to improve language and readability. After using these tools, the authors reviewed and edited the content as needed and take full responsibility for this publication.